\documentclass[sigconf]{acmart}

\usepackage{float}
\usepackage{url}            
\usepackage{subfig}
\usepackage{multirow}
\usepackage{multicol}
\usepackage{amsmath}

\DeclareMathOperator*{\argmin}{arg\,min}

\usepackage{color}

\AtBeginDocument{%
  \providecommand\BibTeX{{%
    \normalfont B\kern-0.5em{\scshape i\kern-0.25em b}\kern-0.8em\TeX}}}

\copyrightyear{2020}
\acmYear{2020}
\setcopyright{acmcopyright}\acmConference[KDD '20]{Proceedings of the 26th ACM SIGKDD Conference on Knowledge Discovery and Data Mining}{August 23--27, 2020}{Virtual Event, CA, USA}
\acmBooktitle{Proceedings of the 26th ACM SIGKDD Conference on Knowledge Discovery and Data Mining (KDD '20), August 23--27, 2020, Virtual Event, CA, USA}
\acmPrice{15.00}
\acmDOI{10.1145/3394486.3403349}
\acmISBN{978-1-4503-7998-4/20/08}

\begin{document}

\fancyhead{}

\title{
Two Sides of the Same Coin:  White-box and Black-box Attacks for Transfer Learning
}

\author{Yinghua Zhang}
\email{yzhangdx@cse.ust.hk}
\affiliation{CSE, HKUST, Hong Kong, China
}

\author{Yangqiu Song}
\email{yqsong@cse.ust.hk}
\affiliation{CSE, HKUST, Hong Kong, China
}
\affiliation{ 
  \institution{Peng Cheng Laboratory, Shenzhen, China}
}

\author{Jian Liang}
\email{joshualiang@tencent.com}
\affiliation{%
  \institution{Cloud and Smart Industries Group, Tencent, China}
}

\author{Kun Bai}
\email{kunbai@tencent.com}
\affiliation{%
  \institution{Cloud and Smart Industries Group, Tencent, China}
}

\author{Qiang Yang}
\email{qyang@cse.ust.hk}
\affiliation{CSE, HKUST, Hong Kong, China
}
\affiliation{%
  \institution{WeBank, China}
}


\renewcommand{\shortauthors}{Zhang, et al.}

\begin{abstract}
  Transfer learning has become a common practice for training deep learning models with limited labeled data in a target domain. On the other hand, deep models are vulnerable to adversarial attacks. Though transfer learning has been widely applied, its effect on model robustness is unclear. To figure out this problem, we conduct extensive empirical evaluations to show that fine-tuning effectively enhances model robustness under white-box FGSM attacks.  
  We also propose a black-box attack method for transfer learning models which attacks the target model with the adversarial examples produced by its source model. To systematically measure the effect of both white-box and black-box attacks, we propose a new metric to evaluate how transferable are the adversarial examples produced by a source model to a target model. Empirical results show that the adversarial examples are more transferable when fine-tuning is used than they are when the two networks are trained independently. 
\end{abstract}



\maketitle

\section{Introduction}
Deep learning models achieve state-of-the-art performances on a wide range of computer vision tasks. Yet the performance is achieved at the cost of large scale labeled training data. In practice, there are many domains where labeled data are insufficient to train a deep model from scratch. In such cases, transfer learning techniques \cite{pan2010survey, weiss2016survey} are usually adopted. Transfer learning uses the knowledge that is extracted from a well-annotated \emph{source} domain to help learning in a \emph{target} domain where only limited labeled data are available. One of the most successful and popular transfer learning techniques is fine-tuning. For example, it is demonstrated that the parameters in a convolutional neural network (CNN) are transferable~\cite{yosinski_nips_2014, oquab_transfer_midlevel_cvpr_2014}. 
Nowadays, there are many pre-trained networks publicly available and developers often use them to save the efforts on data labeling and model training. 

However, a perturbation that is imperceptible to humans can easily fool a deep learning models such as a well-performed complex CNN~\cite{szegedy_iclr_2013}.
A typical example was given in \cite{szegedy_iclr_2013} where a panda image is misclassified as ``gibbon.'' 
Though there are a lot of successful stories of transfer learning, surprisingly, few studies consider the robustness of transfer learning models. 
It is found that adversarial examples can generalize across the networks with different architectures that are trained on the same dataset \cite{liu_arch_transfer_iclr_2016} or the networks that are trained with disjoint datasets \cite{szegedy_iclr_2013}. 
These studies have proven that adversarial examples could be transferable. However, they are not directly dealing with transfer learning models. This motivates us to think about to which extent we can generate adversarial examples for transfer learning.

Generating adversarial examples for transfer learning is not a trivial problem. 
An adversarial attack can be either white-box or black-box. White-box attacks assume that the target of the attack is accessible while black-box attacks only allow querying the network output or even have no knowledge of the network. 
Thus, there are mainly two challenges to be considered. 
First, in the case of white-box attacks, although it has been proved that adversarial training can be transferred to the target domain~\cite{hendrycks_pretrain_robustness_icml_2019}, it is still unclear what is the effect of pure fine-tuning for the resultant model or whether there exists a general way of attacks when only a giant model trained on a large scale dataset (e.g., ImageNet) is available. 
Second, in the case of black-box attacks, to our best knowledge, there has been no study on how transfer learning would affect model robustness under black-box attacks. 
A trivial solution may be directly using the adversarial examples in the source domain that is used for pre-training. However, in many target domains used for fine-tuning, the label sets are different from the source-domain labels. Therefore, it is difficult to apply this trivial solution.

In this paper, we study both white-box and black-box attacks for a simple transfer learning paradigm: the pre-training and fine-tuning procedure of domain adaptation of a CNN model. 
We find this simple transfer learning paradigm shows more robustness under the white-box FGSM attacks. 
For the black-box attack, we propose a simple attack method that attacks the fine-tuned model with the adversarial examples produced by the source model. Experimental results show that this method is simple yet effective and it hurts the robustness results. 
To systematically measure the effect of both white-box and black-box attacks, we propose a new metric to evaluate how transferable are the adversarial examples produced by a source domain network to a target domain network using both white-box and black-box attack results. 
Without loss of generality, we evaluate the following two transfer learning settings.

$\bullet$ The source domain is similar to the target domain. Then we directly transfer the source domain model to the target domain.

$\bullet$ There exists a giant model trained on a general large dataset. However, the similarity of source and target domains does not support to generate adversarial examples for the target domain. In this case, we introduce another source domain which is similar to the target domain and also fine-tuned from the giant model.

\noindent Empirical results show that the adversarial examples are more transferable when fine-tuning is used than they are when the two networks are trained independently. 




In addition to improved transfer performance and robustness under white-box attacks when applying fine-tuning, our study suggests that the benefits are obtained at the cost of the potential risks of using untrusted pre-trained networks. A malicious attacker can take advantage of this phenomenon by releasing a pre-trained model and attack the downstream fine-tuned models. While most transfer learning methods only optimize for a low generalization error, we argue that the robustness of transfer learning models should be considered as well. Otherwise, we may expose transfer learning models under harmful attacks. Such risk has been overlooked which can be dangerous for safety-critical applications such as autonomous driving. 

The rest of this paper is organized as follows. We review related works in Section \ref{sec:related-works}. In Section \ref{sec:problem-setup}, we introduce the problem settings and white-box and black-box attack methods for transfer learning models. The experiment setup is described in Section \ref{sec:experiment-setup}, and numerical results under white-box and black-box attacks are presented in Section \ref{sec:main-results}. Ablation experiments are shown in Section \ref{sec:ablation}. We summarize and discuss the empirical results in Section \ref{sec:discussion}, and finally conclude the paper in Section \ref{sec:conclusion}. 
Our source code is available at \url{https://github.com/HKUST-KnowComp/AttackTransferLearning}.

\section{Related Works}
\label{sec:related-works}

Transfer learning is necessary to overcome the data-hungry nature of neural networks \cite{pan2010survey,weiss2016survey}. Fine-tuning is one of the most popular transfer learning methods. 
Using the networks that are pre-trained on large scale datasets such as ImageNet \cite{russ_imagenet_ijcv_2015} can significantly boost the performance of downstream tasks, such as video classification, object detection, image/video captioning, etc \cite{karpathy_cvpr_2014, girshick_cvpr_2014, vinyals_cvpr_2015, venugopalan_iccv_2015, venugopalan_naacl_2015}. 

In pursuit of machine learning models that are both robust and efficient, adversarial attacks and defenses have attracted attention in the past few years. Numerous attack and defense methods have been proposed \cite{yuan_tnnls_attac_survey_2019}. Szegedy et al. \cite{szegedy_iclr_2013} first propose an L-BFGS method to craft adversarial examples that are close to the original examples and misclassified by the network. L-BFGS attack is effective but slow. To improve efficiency, Goodfellow et al. propose a one-step attack method FGSM by moving along the direction of the gradient \cite{goodfellow_fgsm_iclr_2014}. Madry et al. \cite{madry_pgd_iclr_2018} formulate a min-max optimization problem to study adversarial robustness. They start from a random perturbation around the original input and strengthen the gradient-based attack by applying it iteratively with a small step size. 

\begin{figure*}
    \centering
    \subfloat[\texttt{Scratch}] {\label{fig:transfer_strategy_stratch}
    \includegraphics[width=0.16\textwidth]{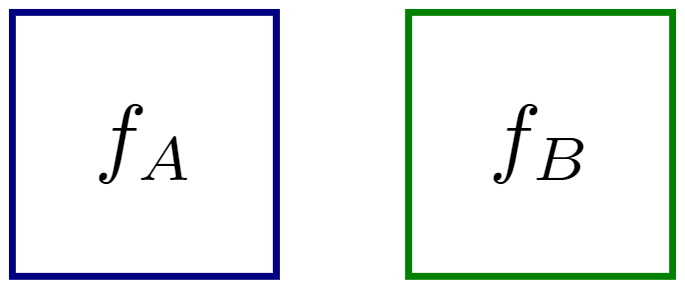}
    }\hspace{0.4in}
    \subfloat[\texttt{FT}] {\label{fig:transfer_strategy_finetune}
    \includegraphics[width=0.20\textwidth]{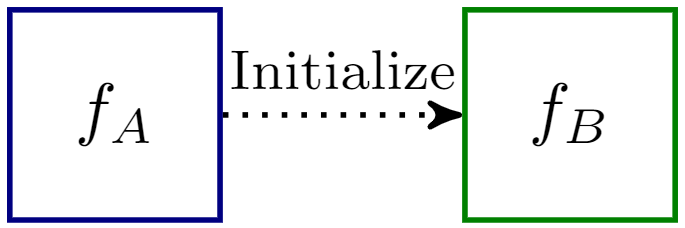}
    }\hspace{0.4in}
    \subfloat[\texttt{CommonInit}] {
    \label{fig:transfer_strategy_common_init}
    \includegraphics[width=0.18\textwidth]{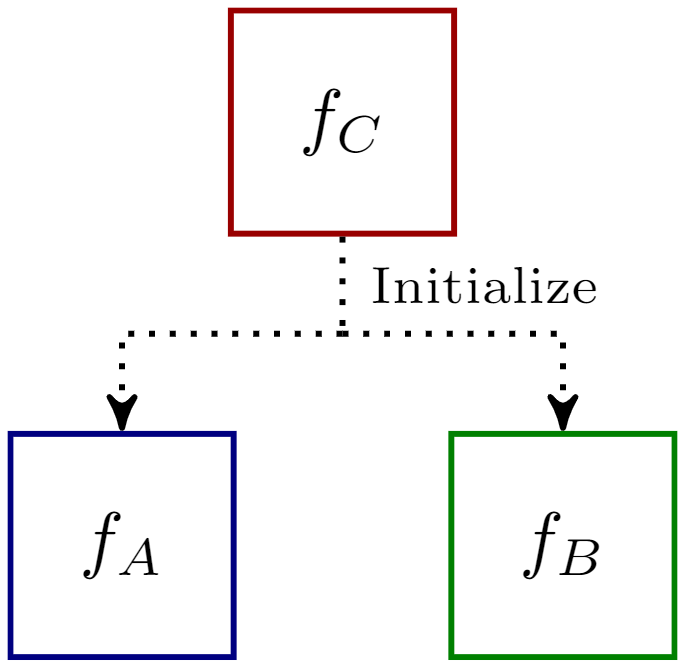}
    }
    \caption{Three model training strategies. There is no transfer learning and \texttt{Model} \texttt{A} and \texttt{Model} \texttt{B} are independent if they are trained with the \texttt{Scratch} strategy. Transfer learning is involved and two models are correlated explicitly/implicitly when the \texttt{FT}/\texttt{CommonInit} strategy is used. }
    \label{fig:transfer_strategies}
\end{figure*}

Previous works that study model robustness assume that the model is trained from scratch in an individual domain while transfer learning techniques are often used in practice and the assumption no longer holds in such settings. Though fine-tuning has been widely used, surprisingly, few research works pay attention to the robustness of transfer learning models. The most related work is \cite{hendrycks_pretrain_robustness_icml_2019} where the robustness of adversarial fine-tuned models under white-box attacks is evaluated. Our work differs with theirs in three aspects: (1) We study the robustness of fine-tuned models under both white-box and black-box attacks; (2) We focus on fine-tuning which is more widely used instead of adversarial fine-tuning proposed in \cite{hendrycks_pretrain_robustness_icml_2019}; (3) Ablation experiments are conducted to study the effect of a number of factors, including the number of labeled data, domain similarity, network architectures, etc., on the transfer performance and robustness. 

\section{Robustness of Transfer Learning Models}
\label{sec:problem-setup}
We first introduce the problem settings and the notations used, then present three model training strategies. We briefly describe how to generate adversarial examples in an individual domain, and then propose the method to attack transfer learning models. We finally introduce a new metric to measure the transferability of adversarial examples between transfer learning models.

\subsection{Problem Setup}
We focus on the classification task where both domains are labeled. To avoid  the confusion with the ``target'' of attack, the source domain and the target domain are called \texttt{Domain} \texttt{A} and \texttt{Domain} \texttt{B}, and denoted by $\mathcal{D}_{A}$ and $\mathcal{D}_{B}$, respectively. \texttt{Domain} \texttt{A} is composed of $n_A$ training samples which is denoted by $\mathcal{D}_{A} = \{(\mathbf{x}_{A_j}, y_{A_j})\}_{j=1}^{n_A}$ where there are $n_A$ training samples $\mathbf{x}_{A} \in \mathcal{X}_{A}$ and their corresponding labels $y_{A} \in \mathcal{Y}_{A}$. Similarly, \texttt{Domain} \texttt{B} is denoted by $\mathcal{D}_{B} = \{(\mathbf{x}_{B_j}, y_{B_j})\}_{j=1}^{n_B}$ and there are $\mathbf{x}_{B} \in \mathcal{X}_{B}$ and $y_{B} \in \mathcal{Y}_{B}$. There are many more labeled data in \texttt{Domain} \texttt{A} than there are in \texttt{Domain} \texttt{B}, i.e., $n_A \gg n_B$. When there are $\mathcal{X}_{A} = \mathcal{X}_{B}$ and $\mathcal{Y}_{A} = \mathcal{Y}_{B}$, the setting is referred to as homogeneous transfer learning. Otherwise, it is referred to as heterogeneous transfer learning. In each domain, a neural network, denoted by $f$, is trained to learn the mapping from the input space to the label space, $f: \mathcal{X} \rightarrow \mathcal{Y}$. The output of the network, denoted by $f(\mathbf{x})$, predicts the probability distribution over the label space. 

\subsection{Model Training Strategies}
\label{sec:model-training-strategies}

As attacking transfer learning can be similar or different from attacking non-transfer learning models, we consider three different training strategies to study the adversarial example generation for comparing transfer learning with non-transfer learning. The three ways to train the models in the two domains, namely \texttt{Scratch}, \texttt{Fine-tune}, and \texttt{CommonInit}, are shown in Fig. \ref{fig:transfer_strategies}. While \texttt{Scratch} is not a transfer learning setting, the other two strategies both involve transfer learning. The \texttt{Fine-tune} and \texttt{CommonInit} strategies address the homogeneous and heterogeneous transfer learning settings, respectively. 
The details of the three training strategies are described as follows.

\noindent$\bullet$ \texttt{Scratch}: As shown in Fig. \ref{fig:transfer_strategy_stratch}, in the \texttt{Scratch} setting, the \texttt{Model} \texttt{B} is randomly initialized and is only trained with \texttt{Domain} \texttt{B}'s data. There is no transfer learning if the models are trained with the \texttt{Scratch} strategy, and \texttt{Model} \texttt{A} and \texttt{Model} \texttt{B} are independent. 

\noindent$\bullet$ \texttt{Fine-tune (FT)}: The \texttt{FT} strategy is shown in Fig. \ref{fig:transfer_strategy_finetune}. To transfer the parameters from \texttt{Domain} \texttt{A} to \texttt{Domain} \texttt{B}, the two networks share an identical architecture. \texttt{Model} \texttt{A} ($f_{A}$) is first trained with \texttt{Domain} \texttt{A}'s data, then \texttt{Model} \texttt{B} is initialized with the parameters of \texttt{Model} \texttt{A}. Finally, \texttt{Model} \texttt{B} ($f_{B}$) is fine-tuned with \texttt{Domain} \texttt{B}'s data. 

\noindent$\bullet$ \texttt{CommonInit}: Both \texttt{Model} \texttt{A} and \texttt{Model} \texttt{B} are initialized with another \texttt{Model} \texttt{C} and then fine-tuned with domain-specific data. This approach is useful for black-box attacking a heterogeneous transfer learning model. 
For example, to train a model on \texttt{STL10} where there are only $5,000$ training images, a natural choice of the source domain is a downsampled variant of ImageNet \cite{chrabaszcz_imagenet32_arxiv_2017}, denoted by \texttt{ImageNet32}, where there are more than one million training images. However, the label spaces of the two domains do not agree. There are $10$ classes and $1,000$ classes in \texttt{STL10} and \texttt{ImageNet32}, respectively. We cannot attack an \texttt{STL10} model with the adversarial examples produced by an \texttt{ImageNet32} based model due to the mismatched label space. One solution to the problem is to use another domain such as \texttt{CIFAR10} as the source domain. \texttt{CIFAR10} has the same label space as \texttt{STL10} does. Thus, the adversarial examples generated based on \texttt{CIFAR10} can be transferred to attack models based on \texttt{STL10}. In our setting, both models for \texttt{CIFAR10} and \texttt{STL10} are fine-tuned from \texttt{ImageNet32} and the adversarial examples produced by the \texttt{CIFAR10} model can be more transferable to attack the \texttt{STL10} based model. In this example, \texttt{CIFAR10}, \texttt{STL10}, and \texttt{ImageNet32} correspond to \texttt{Domain} \texttt{A}, \texttt{B}, and \texttt{C} in Fig. \ref{fig:transfer_strategy_common_init}, respectively.

\subsection{Generate Adversarial Examples}
Before attacking transfer learning models, we first introduce adversarial example generation in an individual domain as preliminary knowledge. There are different ways to generate adversarial examples. As our study mainly focuses on different transfer learning settings and study the transferability of the generated examples in different settings, we choose the widely used Fast Sign Gradient Method (FGSM) \cite{goodfellow_fgsm_iclr_2014} in our work. 

In general, crafting adversarial examples for a model $f$ can be formulated as an optimization problem: 
\begin{equation}
    \argmin_{\lVert \mathbf{\hat{x}} - \mathbf{x} \rVert_{p} \leq \epsilon} \ell (\hat{y}, f(\mathbf{\hat{x}})), 
    \label{eqn:adv_example_opt_problem}
\end{equation}
where $\mathbf{\hat{x}}$ denotes the adversarial example, $\hat{y}$ denotes a label that is different from the ground truth label $y$, $\ell(\cdot, \cdot)$ denotes a classification loss, $\lVert \cdot \rVert_p$ denotes the $p$-norm distance and $\epsilon$ denotes the perturbation budget. The adversarial example is optimized to mislead the network $f$ within the $p$-norm $\epsilon$-ball of the clean example $\mathbf{x}$. In this paper, we adopt the cross-entropy loss as the classification loss and the infinity norm as the distance measure. 

Many methods to solve the optimization problem in Eq. (\ref{eqn:adv_example_opt_problem}) have been proposed \cite{szegedy_iclr_2013,goodfellow_fgsm_iclr_2014,madry_pgd_iclr_2018}. The FGSM takes one step in the direction of the gradient: 
\begin{equation}
    \mathbf{\hat{x}} = \mathbf{x} + \epsilon \cdot sgn(\nabla_{\mathbf{x}} \ell (y, f(\mathbf{x}))), 
\end{equation}
where the $sgn$ function extracts the sign of each dimention in the gradient $\nabla_{\mathbf{x}} \ell (y, f(\mathbf{x}))$ and uses that as the direction to slightly modify the given example. The FGSM update is believed to optimize Eq. (\ref{eqn:adv_example_opt_problem}) to generate some valid examples that are imperceptible to humans but may fool a deep learning model  \cite{goodfellow_fgsm_iclr_2014}. 

\begin{figure}
    \centering
    \includegraphics[width=0.45\textwidth]{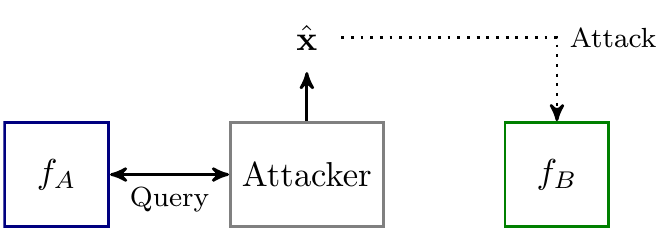}
    \caption{Black-box attack \texttt{Model} \texttt{B} with adversarial examples produced by \texttt{Model} \texttt{A}. The proposed method allows attack without any query to the target model. }
    \label{fig:blackbox-attack-by-transfer}
\end{figure}

\subsection{Attack Transfer Learning Models}
We consider the robustness of transfer learning models under both white-box and black-box attacks in this paper. Particularly, we apply the FGSM attack method to generate adversarial examples for transfer learning models under both white-box and black-box settings.

\subsubsection{White-box Attacks}
For white-box attacks, it is assumed that everything related to the target of the attack is accessible. Thus, we can view \texttt{Model} \texttt{B} as a white box and attack it by applying FGSM. \texttt{Model} \texttt{B} is either trained from scratch in \texttt{Domain} \texttt{B} or fine-tuned from a source network \texttt{Model} \texttt{A}. 

\subsubsection{Black-box Attacks}
The assumption of white-box attacks usually does not hold in reality. A more realistic setting is the \emph{black-box} attack where the target of the attack is completely not accessible or only the output of the target can be queried. While most black-box attack methods are developed for the setting that allows querying the output of the target model \cite{papernot_substitute_acm_ccs_2017, chen_zoo_acm_ais_2017}, we focus on a more restricted setting where no access to the target model is allowed. 

We develop a simple black-box attack method for transfer learning models which first produces an adversarial example with \texttt{Model} \texttt{A} and then attacks \texttt{Model} \texttt{B} with the generated adversarial example. Hence we can attack \texttt{Model} \texttt{B} without any access to it. The procedure is illustrated in Fig. \ref{fig:blackbox-attack-by-transfer}. The two models can be trained with three different strategies introduced in Section \ref{sec:model-training-strategies}. We will apply the proposed black-box attack method and evaluate model robustness in Section \ref{sec:blackbox-robustness}. 


\subsection{Transferability of Adversarial Examples}

Here, we propose a new metric to measure the transferability of adversarial examples between transfer learning models.
Similar to the evaluation metrics for the robustness under white-box attacks, the robustness under black-box attacks can be measured by the adversarial accuracy. A lower adversarial accuracy indicates that the model is more vulnerable to the transferred adversarial examples. 

As will be shown in Section \ref{sec:white-box-robustness}, the model robustness can be enhanced after fine-tuning, and hence it may be unfair to directly compare the adversarial accuracy of the \texttt{Scratch} model and that of a transfer learning model. We introduce a new metric to evaluate the transferability of adversarial examples between \texttt{Model} \texttt{A} and \texttt{Model} \texttt{B}. Let $a_{w}$ and $a_{b}$ denote the adversarial accuracy of a network under the white-box attack and black-box attack, respectively. Let $\gamma$ denote the transferability metric defined as:
\begin{equation}
    \gamma = \frac{a_{b} - a_{w}}{a_{w}}, 
\end{equation}
which measures how much the adversarial accuracy under the black-box attack deviates from the one under the white-box attack. Usually, we have $\gamma > 0$ since less knowledge is available in the black-box setting and black-box attacks are not as effective as the white-box ones. If $\gamma \leq 0$, it means that the black-box attacks by transfer learning are even more effective than directly attacking the target model. 

{
\begin{table}[t]
    \centering
    \begin{tabular}{cccccc}
    \toprule
        \multicolumn{3}{c}{Domain} & \multirow{2}{*}{$n_A$} & \multirow{2}{*}{$n_B$} & \multirow{2}{*}{$n_C$} \\
        \cline{1-3}
        \texttt{A} & \texttt{B} & \texttt{C} &  &  &  \\
        \midrule
        \texttt{M} & \texttt{U} & \texttt{S} & $60K$ & $74$  & $604K$ \\
        \texttt{U} & \texttt{M} & \texttt{S} & $7.4K$ & $600$ & $604K$ \\
        \texttt{S} & \texttt{M} & NA & $604K$ & $600$ & NA \\
        \texttt{S} & \texttt{Syn} & \texttt{M} & $604K$ & $4.8K$ & $60K$ \\
        \texttt{CIFAR} & \texttt{STL} & \texttt{ImageNet32} & $45K$ & $4.5K$ &  $1.28M$ \\
        \bottomrule
    \end{tabular}
    \caption{Statistics of transfer tasks.}
    \label{tab:transfer_tasks}
\end{table}
}
\section{Experiment Setup}
\label{sec:experiment-setup}
In this section, we introduce the datasets, evaluation metrics and implementation details in the following. 
\subsection{Transfer Tasks}
We use seven datasets for our evaluation, which are \texttt{MNIST} \texttt{(M)} \cite{lecun_mnist_ieee_proc_1998}, \texttt{USPS} \texttt{(U)} \cite{hull_usps_TPAMI_1994}, \texttt{SVHN} \texttt{(S)} \cite{netzer_svhn_tr_2011}, \texttt{SynDigits} \texttt{(Syn)} \cite{ganin_dann_jmlr_2016}, \texttt{CIFAR10} \cite{krizhevsky_cifar_tr_2009}, \texttt{STL10} \cite{coates_stl_aistat_2011}, and \texttt{ImageNet32} \cite{chrabaszcz_imagenet32_arxiv_2017}. The first four datasets contain ``0'' to ``9'' digit images with various distributions. Both \texttt{M} and \texttt{U} are handwritten digit databases while \texttt{S} and \texttt{Syn} are digit images with colored backgrounds. The latter three datasets are composed of low-resolution natural images. 
Five transfer tasks in the form of (\texttt{Domain} \texttt{A}, \texttt{Domain} \texttt{B}, \texttt{Domain} \texttt{C}) are constructed\footnote{If $n_B$ is smaller than the size of the default training set, we randomly sample $n_B$ examples from it. Usually, there are a large amount of data in \texttt{Domain} \texttt{C}. Since there are only $7.4K$ training examples in \texttt{U}, we do not use \texttt{U} as \texttt{Domain} \texttt{C}. For the \texttt{CIFAR} $\to$ \texttt{STL} task, 9 categories that are shared by the two domains are used. }. Their statistics are described in Table \ref{tab:transfer_tasks}. We follow the default train/test split of the datasets. 
As preprocessing, all the images are resized to $32 \times 32$ and they are rescaled to the range $[-1, 1]$.

\begin{figure*}
    \subfloat[Classification accuracy]{
    \begin{minipage}{0.3\linewidth}
    \centering
    \includegraphics[scale=0.4]{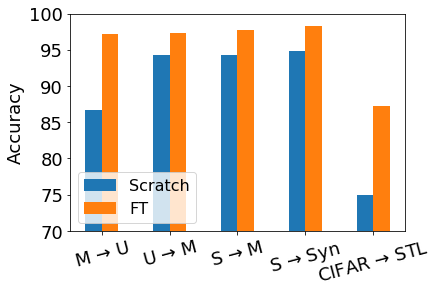}
    \label{fig:clean_accuracy}
    \end{minipage}
    }
    \hfill
    \subfloat[\texttt{M} $\to$ \texttt{U}] {
    \begin{minipage}{0.3\linewidth}
    \centering
    \includegraphics[scale=0.4]{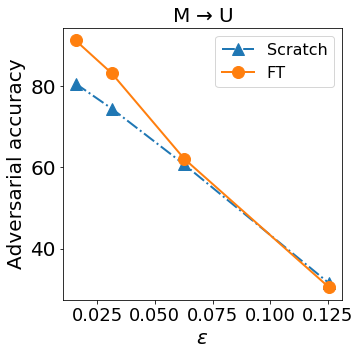}
    \end{minipage}
    }
    \hfill
    \subfloat[\texttt{U} $\to$ \texttt{M}] {
    \begin{minipage}{0.3\linewidth}
    \centering
    \includegraphics[scale=0.4]{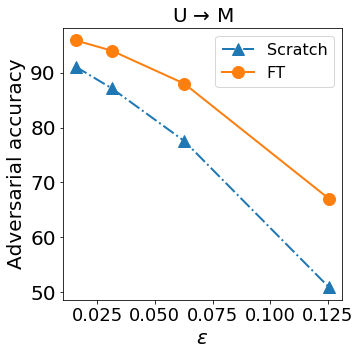}
    \end{minipage}
    }
    \\
    \subfloat[\texttt{S} $\to$ \texttt{M}] {
    \begin{minipage}{0.3\linewidth}
    \centering
    \includegraphics[scale=0.4]{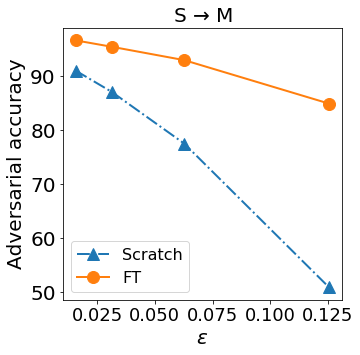}\label{fig:non-targeted-white-box_c}
    \end{minipage}
    }
    \hfill
    \subfloat[\texttt{S} $\to$ \texttt{Syn}] {
    \begin{minipage}{0.3\linewidth}
    \centering
    \includegraphics[scale=0.4]{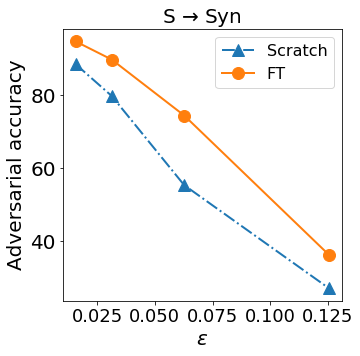}
    \end{minipage}
    }
    \hfill
    \subfloat[\texttt{CIFAR} $\to$ \texttt{STL}] {
    \begin{minipage}{0.3\linewidth}
    \centering
    \includegraphics[scale=0.4]{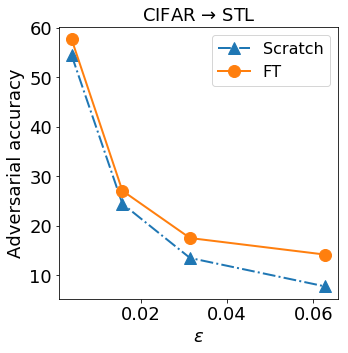}
    \end{minipage}
    }
    \caption{(a) Classification accuracy of the five transfer tasks. The \texttt{FT} models consistently outperform the \texttt{Scratch} baselines. (b-f) Robustness under white-box FGSM attacks. Compared to the \texttt{Scratch} models, the adversarial accuracy increases after fine-tuning, which indicates that the \texttt{FT} models are more robust than the \texttt{Scratch} ones. }
    \label{fig:non-targeted-white-box}
\end{figure*}

\subsection{Evaluation Metrics}
Since the five transfer tasks are all classification tasks, the classification accuracy on clean examples is adopted to measure the transfer performance in \texttt{Domain} \texttt{B}. A higher classification accuracy indicates better transfer performance. 
In addition to the transfer performance, robustness under adversarial attacks is considered as well. The robustness of a neural network is measured by the classification accuracy on the adversarial examples, which is referred to as \emph{adversarial accuracy} in the following. The adversarial examples are obtained on the test set of \texttt{Domain} \texttt{B}. The clean examples that are correctly predicted by the target model are attacked. The higher the adversarial accuracy is, the more robust the network is. 

\subsection{Implementation Details}
All the experiments are implemented with the {PyTorch} deep learning framework. Two network architectures are adopted. For the digit classification tasks, a simple $5$-layer CNN, denoted by \texttt{DTN}, is used. For the \texttt{CIFAR} $\to$ \texttt{STL} task, a more expressive architecture, the 28-10 wide residual network (\texttt{WideRes}) \cite{zagoruyko_wrn_arxiv_2016}, is used. When \texttt{ImageNet32} is used as \texttt{Domain} \texttt{C}, \texttt{Model} \texttt{A} and \texttt{B} are initialized with \texttt{Model} \texttt{C} except for the final classification layer. The neural networks are optimized with the mini-batch stochastic gradient descent with the momentum of 0.9. Early stopping is used, that is, if the network performance does not improve within 50 epochs, the training process is terminated. The batch size equals to 128. The learning rate is selected from $\{0.1, 0.01, 10^{-3}\}$ and weight decay is selected from $\{5\times 10^{-4}, 2.5 \times 10^{-5}, 5\times 10^{-6}\}$. We report the accuracy achieved with the optimal hyperparameters. 

\section{Main Results}
\label{sec:main-results}
In this section, we study the effect of transfer learning on the robustness of \texttt{Model} \texttt{B} under attacks. 

\subsection{Under White-box Attacks}
\label{sec:white-box-robustness}


The classification results of the five transfer tasks are shown in Fig. \ref{fig:clean_accuracy}. On all transfer tasks, fine-tuning brings noticeable improvement over the \texttt{Scratch} baselines, which demonstrates the necessity and effectiveness of transfer learning. 
The adversarial accuracy of the five transfer tasks are shown in Fig. \ref{fig:non-targeted-white-box}. We report adversarial accuracies under multiple perturbation budgets. Compared to the adversarial accuracy of \texttt{Scratch} models, the adversarial accuracy increases after fine-tuning. The improvement is more obvious when the network is attacked with a large perturbation budget. For example, the adversarial accuracy rises from $50.86\%$ to $84.96\%$ on the \texttt{S} $\to$ \texttt{M} task when $\epsilon=0.125$. The results show that in addition to better transfer performance, another advantage of fine-tuning is the enhanced model robustness under white-box attacks. 

\begin{figure*}
    \subfloat[Domain (A, B, C) = (\texttt{M}, \texttt{U}, \texttt{S})] {
    \begin{minipage}{0.24\linewidth}
    \centering
    \includegraphics[scale=0.35]{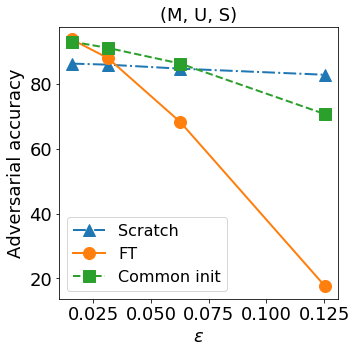}
    \end{minipage}
    }
    \hfill
    \subfloat[Domain (A, B, C) = (\texttt{U}, \texttt{M}, \texttt{S})] {
    \begin{minipage}{0.24\linewidth}
    \centering
    \includegraphics[scale=0.35]{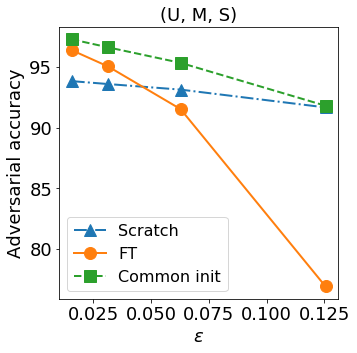}
    \end{minipage}
    }
    \hfill
    \subfloat[Domain (A, B, C) = (\texttt{S}, \texttt{Syn}, \texttt{M})] {
    \begin{minipage}{0.24\linewidth}
    \centering
    \includegraphics[scale=0.35]{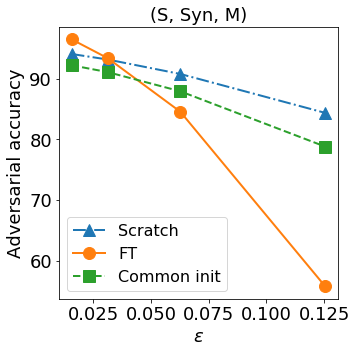}\label{fig:non-targeted-abs-black-box_c}\label{fig:non-targeted-abs-black-box_c}
    \end{minipage}
    }
    \hfill
    \subfloat[Domain (A, B, C) = (\texttt{CIFAR}, \texttt{STL}, \texttt{ImageNet32})] {
    \begin{minipage}{0.24\linewidth}
    \centering
    \includegraphics[scale=0.35]{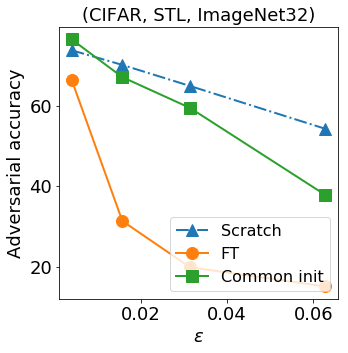}
    \end{minipage}
    }
    \caption{Robustness (adversarial accuracy) under black-box attacks. The adversarial accuracies of the \texttt{FT} and \texttt{CommonInit} models drop drastically when the perturbation budget $\epsilon$ increases. They are much lower than those obtained with the \texttt{Scratch} models, which indicates that the fine-tuned models are likely to be attacked by the adversarial examples produced by their source models. }
    \label{fig:non-targeted-abs-black-box}
\end{figure*}
\begin{figure*}
    \subfloat[Domain (A, B, C) = (\texttt{M}, \texttt{U}, \texttt{S})] {
    \begin{minipage}{0.24\linewidth}
    \centering
    \includegraphics[scale=0.35]{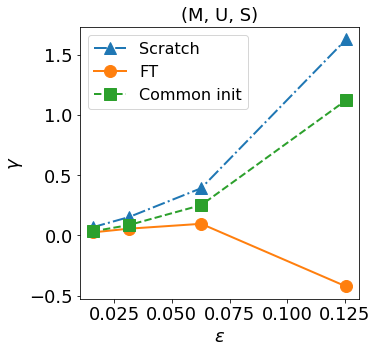}
    \end{minipage}
    }
    \hfill
    \subfloat[Domain (A, B, C) = (\texttt{U}, \texttt{M}, \texttt{S})] {
    \begin{minipage}{0.24\linewidth}
    \centering
    \includegraphics[scale=0.35]{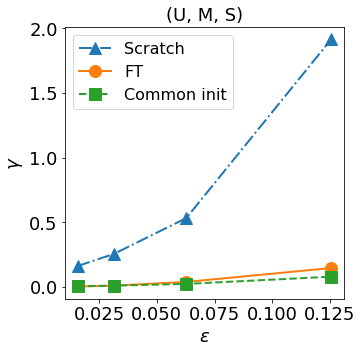}
    \end{minipage}
    }
    \hfill
    \subfloat[Domain (A, B, C) = (\texttt{S}, \texttt{Syn}, \texttt{M})] {
    \begin{minipage}{0.24\linewidth}
    \centering
    \includegraphics[scale=0.35]{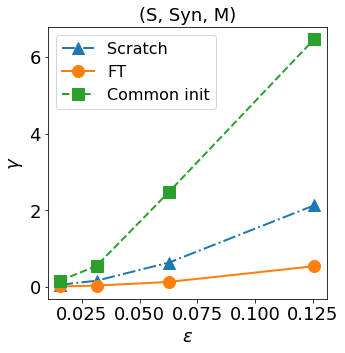}\label{fig:non-targeted-gamma-black-box_c}
    \end{minipage}
    }
    \hfill
    \subfloat[Domain (A, B, C) = (\texttt{CIFAR}, \texttt{STL}, \texttt{ImageNet32})] {
    \begin{minipage}{0.24\linewidth}
    \centering
    \includegraphics[scale=0.35]{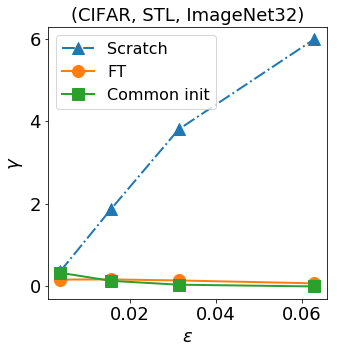}
    \end{minipage}
    }
    \caption{Robustness (transferability $\gamma$) under black-box attacks. The new metric $\gamma$ considers both white-box and black-box attack results and evaluates how transferable are the adversarial examples produced by \texttt{Model} \texttt{A} to \texttt{Model} \texttt{B}. When $\gamma$ drops below $0$, it means that \texttt{Model} \texttt{B} is more vulnerable to the adversarial examples transferred from \texttt{Model} \texttt{A} than those crafted directly with \texttt{Model} \texttt{B} as a white-box.}
    \label{fig:non-targeted-gamma-black-box}
\end{figure*}

\subsection{Under Black-box Attacks}
\label{sec:blackbox-robustness}
The adversarial accuracy and the transferability of the four transfer tasks are shown in Figs. \ref{fig:non-targeted-abs-black-box} and \ref{fig:non-targeted-gamma-black-box}. In terms of the absolute adversarial accuracy values, the adversarial accuracy drops as the perturbation budget $\epsilon$ increases. While the adversarial accuracy of the \texttt{Scratch} model remains rather stable under multiple $\epsilon$ values, the adversarial accuracies of \texttt{CommonInit} and \texttt{FT} models drop significantly. Consequently, when the perturbation budget $\epsilon$ is large, the adversarial accuracies of \texttt{CommonInit} and \texttt{FT} models are much lower than those of the \texttt{Scratch} model. When there is Domain (A, B, C) = (\texttt{M}, \texttt{U}, \texttt{S}), the adversarial accuracy of the \texttt{Scratch} model remains larger than $80\%$ while the adversarial accuracies of the \texttt{CommonInit} and \texttt{FT} model are only $70.59\%$ and $17.53\%$ when $\epsilon = 0.125$, respectively.

When the model robustness is measured by the transferability $\gamma$, the $\gamma$ values of the \texttt{CommonInit} and \texttt{FT} models are usually smaller than those of the \texttt{Scratch} model, which indicates that \texttt{Model} \texttt{B} is likely to be successfully attacked by the adversarial examples produced by \texttt{Model} \texttt{A} if the parameters of the two models are correlated either explicitly or implicitly. An exception is found on the 
Domain (A, B, C) = (\texttt{S}, \texttt{Syn}, \texttt{M})
task where the $\gamma$ values of the \texttt{CommonInit} model are larger than those of the \texttt{Scratch} model. We hypothesize that this is because the source domain \texttt{M} which is composed of handwritten digits is quite different from the two target domains \texttt{S} and \texttt{Syn}. The $\gamma$ value of the \texttt{FT} model drops below $0$ on the 
Domain (A, B, C) = (\texttt{M}, \texttt{U}, \texttt{S})
task when $\epsilon = 0.125$, which means that \texttt{Model} \texttt{B} is more vulnerable to the adversarial examples transferred from \texttt{Model} \texttt{A} than those crafted directly with \texttt{Model} \texttt{B} as a white-box. The results of the black-box attacks show that fine-tuning might introduce potential risks of being attacked by its source model, which are unaware of previously. 
\begin{figure*}[t]
    \centering
    \subfloat[Transfer performance] {
    \begin{minipage}{0.48\linewidth}
    \centering
    \includegraphics[scale=0.3]{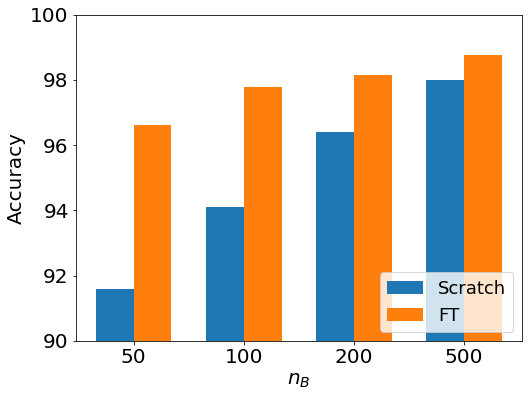}
    \end{minipage}
    }
    \hfill
    \subfloat[Under white-box attacks] {
    \begin{minipage}{0.48\linewidth}
    \centering
    \includegraphics[scale=0.3]{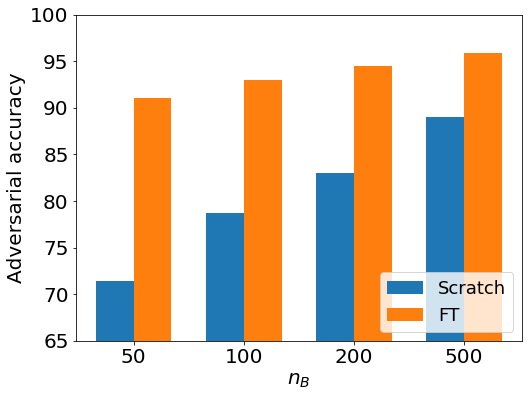}
    \end{minipage}
    }
    \vspace{-0.1in}
    \caption{The effect of $n_B$. As more labeled data are available in \texttt{Domain} \texttt{B}, both the transfer performance and the model robustness are improved. 
   We report the adversarial accuracy of the white-box attacks with $\epsilon = 16/255$ in (b).
    }
    \vspace{-0.2in}
    \label{fig:ablation_nb}
\end{figure*}

\begin{figure*}[t]
    \centering
    \subfloat[Transfer performance] {
    \begin{minipage}{0.48\linewidth}
    \centering
    \includegraphics[scale=0.3]{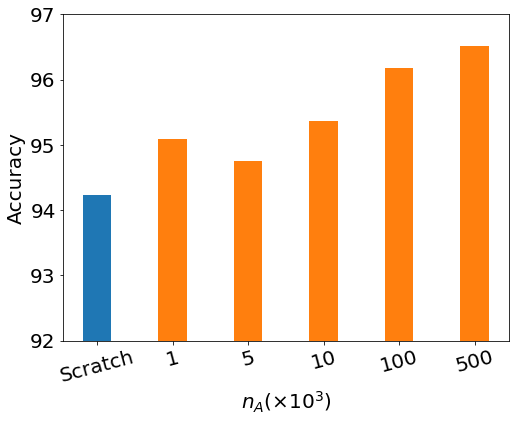}
    \end{minipage}
    }
    \hfill
    \subfloat[Under white-box attacks] {
    \begin{minipage}{0.48\linewidth}
    \centering
    \includegraphics[scale=0.3]{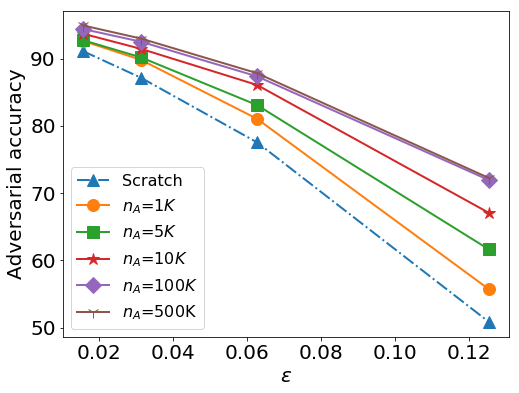}
    \end{minipage}
    }
    \vspace{-0.1in}
    \caption{The effect of $n_A$. The \texttt{FT} models consistently outstrip the \texttt{Scratch} model. When $n_A$ increases, both the transfer performance and the model robustness are improved. }
    \label{fig:ablation_na}
    \vspace{-0.2in}
\end{figure*}

\section{Ablations}
\label{sec:ablation}
There are a number of factors that might affect the transfer performance and model robustness. To provide further insights into the effect of transfer learning, we conduct ablation experiments to study the following factors: (1) The number of training examples in \texttt{Domain} \texttt{B}. 
(2) The number of training examples in \texttt{Domain} \texttt{A}. 
(3) The choice of \texttt{Domain} \texttt{A}. 
(4) Network architectures. 
The ablation experiments are conducted on the 
\texttt{S} $\to$ \texttt{M} 
task and the network architecture is \texttt{DTN} if not specified. 

\subsection{Number of Training Examples in \texttt{Domain} \texttt{B}}
\label{sec:ablation_nb}
We first show the impact of varying $n_B$ on the transfer performance and model robustness in Fig. \ref{fig:ablation_nb}. The \texttt{Scratch} models and \texttt{FT} models are trained with $50, 100, 200, 500$ labeled samples in \texttt{Domain} \texttt{B}. We report the adversarial accuracy when  $\epsilon = 16/255$. The classification accuracy and the adversarial accuracy of both models increase as more labeled data are available. The transfer performance and the model robustness of the \texttt{FT} model remain stable given multiple values of $n_B$ while those of the \texttt{Scratch} model significantly drops as $n_B$ decreases. For example, when there are only $50$ examples in \texttt{Domain} \texttt{B}, the classification accuracy and the adversarial accuracy of the \texttt{Scratch} model is $91.59\%$ and $71.45\%$, respectively, which lags behind those of the \texttt{FT} model by $4.99\%$ and $14.04\%$. The results show that fine-tuning can be particularly beneficial when there are very few labeled samples in \texttt{Domain} \texttt{B}. 

\subsection{Number of Training Examples in \texttt{Domain} \texttt{A}}
\label{sec:ablation_na}
We vary the number of training examples in \texttt{Domain} \texttt{A} and report the classification accuracy and adversarial accuracy in Fig. \ref{fig:ablation_na}. \texttt{Model} \texttt{A} is trained with $1K, 5K, 10K, 100K$ and $500K$ examples. Similar to the results in Section \ref{sec:ablation_nb}, more labeled data in \texttt{Domain} \texttt{A} yield better transfer performance and improved robustness. The \texttt{FT} models obtained with different $n_A$ values consistently outperforms the \texttt{Scratch} model. 

\subsection{Choice of \texttt{Domain} \texttt{A}}
\begin{figure*}[t]
    \centering
    \subfloat[Transfer performance] {
    \includegraphics[scale=0.225]{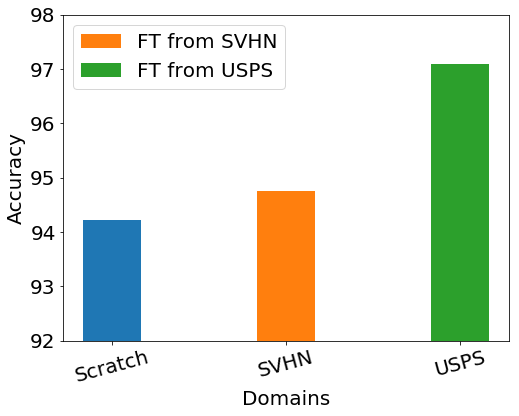}
    }
    \subfloat[Under white-box attacks] {
    \includegraphics[scale=0.225]{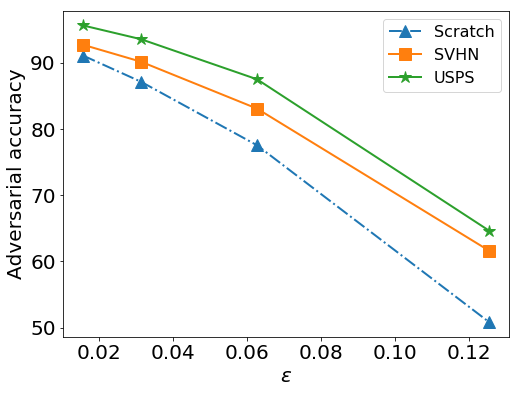}
    }
    \subfloat[Under black-box attacks] {
    \includegraphics[scale=0.225]{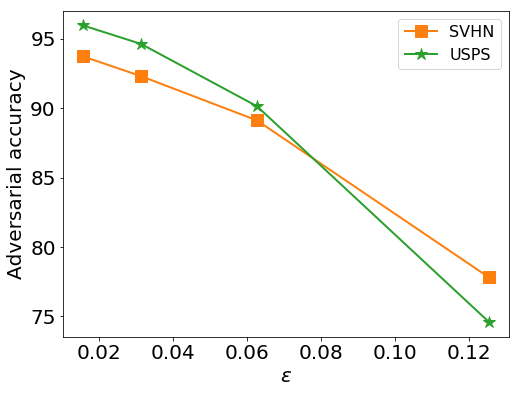}
    }
    \subfloat[Transferability $\gamma$] {
    \includegraphics[scale=0.225]{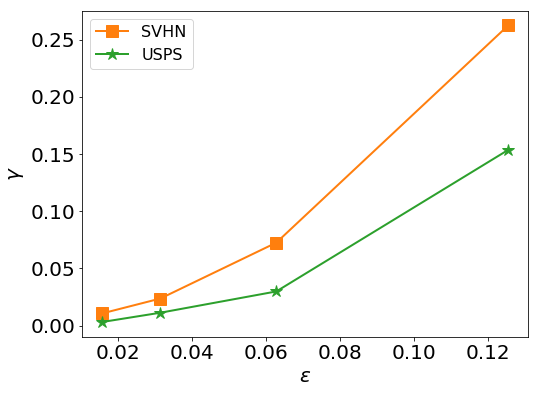}
    }
    \caption{
    The effect of the choice of \texttt{Domain} \texttt{A} on transfer performance and robustness: \texttt{S} $\to$ \texttt{M} vs. \texttt{U} $\to$ \texttt{M}. \texttt{U} is more similar to \texttt{M} compared to \texttt{S}. Fine-tuning from a more visually similar domain yields more performance gain under white-box attacks, but the adversarial examples are more transferable if they are produced by the source model that is trained on a more similar domain. 
    }
    \label{fig:ablation_domain_similarity}
\end{figure*}

To study the effect of different source domains, we fix \texttt{M} as the \texttt{Domain} \texttt{B} and use \texttt{S} and \texttt{U} as \texttt{Domain} \texttt{A}, respectively. The number of training samples in \texttt{Domain} \texttt{A} and \texttt{Domain} \texttt{B} are $5,000$ and $600$, respectively. Thus we have two transfer tasks, \texttt{S} $\to$ \texttt{M} and \texttt{U} $\to$ \texttt{M}. Both \texttt{M} and \texttt{U} are handwritten digits and hence they
are more visually similar than the other task
\texttt{S} $\to$ \texttt{M}. 
The results are shown in Fig. \ref{fig:ablation_domain_similarity}. With the same amount of labeled data, when \texttt{Domain} \texttt{A} is more similar to \texttt{Domain} \texttt{B}, the classification accuracy is higher and \texttt{Model} \texttt{B} is more robust under white-box attacks. At the same time, \texttt{Model} \texttt{B} is more vulnerable to adversarial examples transferred from \texttt{Model} \texttt{A}. 

\begin{figure*}[t]
    \centering
    \subfloat[Transfer performance] {
    \includegraphics[scale=0.235]{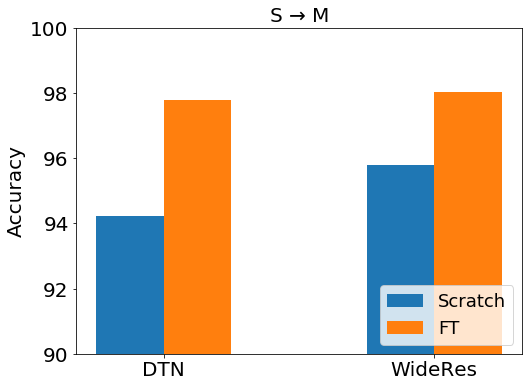}\label{fig:ablation_arch_a}
    }
    \subfloat[Under white-box attacks] {
    \includegraphics[scale=0.225]{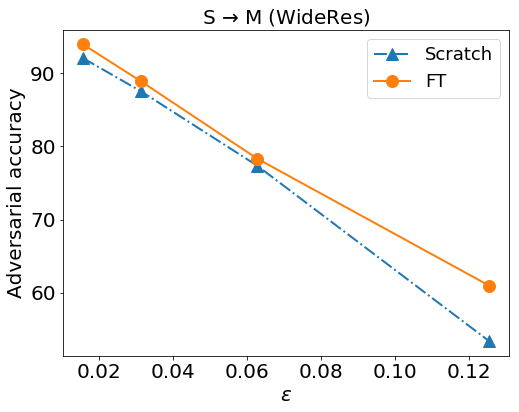}\label{fig:ablation_arch_b}
    }
    \subfloat[Under black-box attacks] {
    \includegraphics[scale=0.225]{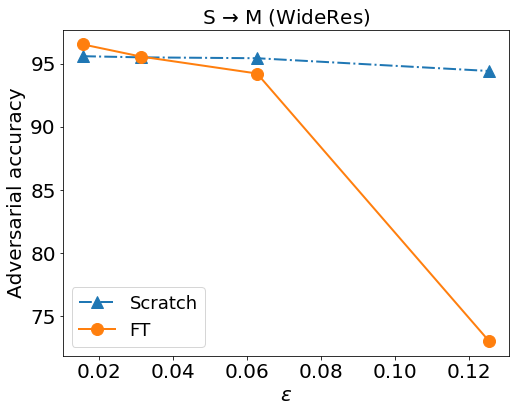}\label{fig:ablation_arch_c}
    }
    \subfloat[Transferability $\gamma$] {
    \includegraphics[scale=0.225]{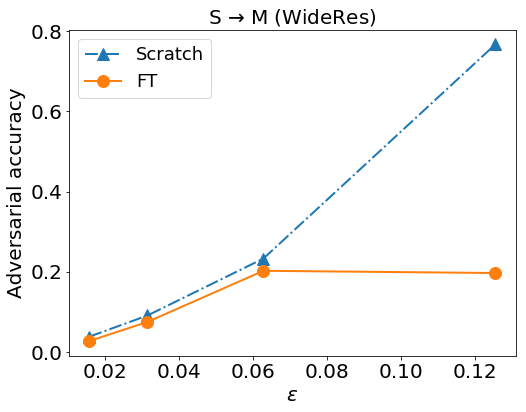}\label{fig:ablation_arch_d}
    }
    \caption{
    The effect of network architectures on transfer performance and robustness. 
    The results of \texttt{WideRes} are consistent with those of the \texttt{DTN} architecture. The \texttt{FT} model is more advantageous than the \texttt{Scratch} model when the \texttt{WideRes} architecture is adopted under white-box attacks. On the other hand, the \texttt{FT} model is more likely to be attacked by the adversarial examples produced by its source model than the \texttt{Scratch} model under black-box attacks. 
    }
    \label{fig:ablation_arch}
\end{figure*}

\subsection{Network Architectures}
We examine whether the observations generalize to other network architectures. We repeat the experiment with the \texttt{WideRes} network and the results are shown in Fig. \ref{fig:ablation_arch}. \texttt{WideRes} networks use widened residual blocks that improve both model performance and training efficiency. There are $28$ convolutional layers in a \texttt{WideRes} network while there are only $5$ layers in a \texttt{DTN} network, and the \texttt{WideRes} network has more representational power. This is demonstrated by the fact that the classification accuracy is improved for both the \texttt{Scratch} and \texttt{FT} model (Fig. \ref{fig:ablation_arch_a}). Moreover, the adversarial accuracy of the \texttt{FT} model consistently outperforms that of the \texttt{Scratch} model disregard of the network architecture that is used. 
In terms of the robustness under black-box attacks, 
the conclusions are the same as those drawn when the \texttt{DTN} architecture is used: 
the \texttt{FT} model is again more likely to be attacked by the adversarial examples produced by its source model than the \texttt{Scratch} model.

\begin{figure*}[t]
    \centering
    \subfloat[Cumulative histograms of gradient norms] {
    \begin{minipage}{0.3\linewidth}
    \centering
    \includegraphics[scale=0.3]{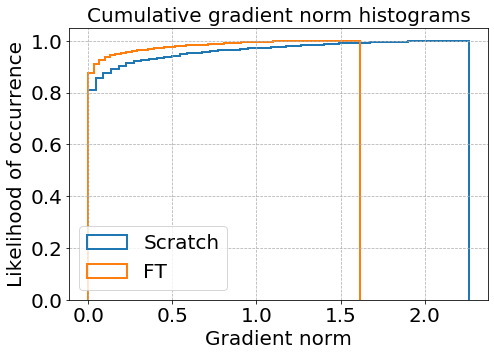}
    \label{fig:chist_gn}
    \end{minipage}
    }
    \hfill
    \subfloat[Histogram of gradient norms of the \texttt{Scratch} model] {
    \begin{minipage}{0.3\linewidth}
    \centering
    \includegraphics[scale=0.3]{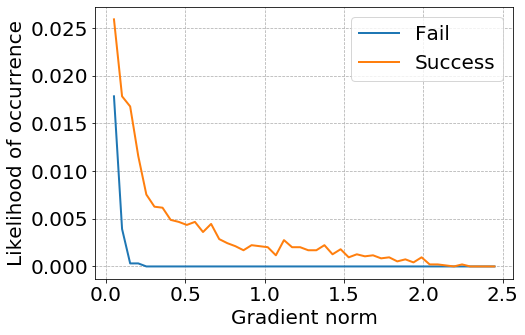}
    \label{fig:scratch_hist_gn}
    \end{minipage}
    }
    \hfill
    \subfloat[Histogram of gradient norms of the \texttt{FT} model] {
    \begin{minipage}{0.3\linewidth}
    \centering
    \includegraphics[scale=0.3]{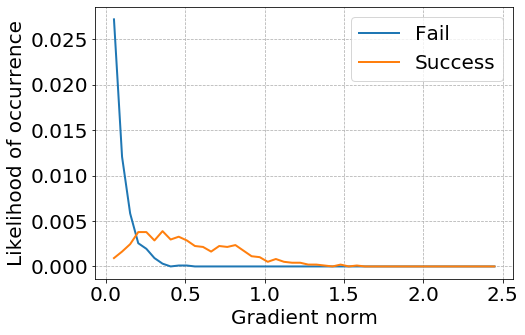}
    \label{fig:ft_hist_gn}
    \end{minipage}
    }
    \caption{The histograms of the gradient norms. The \texttt{FT} model is more likely to produce small gradient norms, which indicates an improved Lipschitzness of the loss landscape and hence makes the model more robust. (For better viewing experience, the gradient norm starts from $0.05$ in (b) and (c). )
    }
    \label{fig:ablation_grad_norms}
\end{figure*}

\section{Discussion}
\label{sec:discussion}
As demonstrated by the experimental results and ablation studies, fine-tuning can effectively improve both transfer performance and robustness under white-box attacks. On the other hand, we can successfully attack a fine-tuned target model in a restricted black-box manner by utilizing its source model, which is a downside of fine-tuning. The observations generalize across different transfer tasks and network architectures. We hypothesize that the success of fine-tuning can be attributed to the following two reasons.

$\bullet$ The improvements may benefit from more training data. Our empirical discoveries in Sections \ref{sec:ablation_nb} and \ref{sec:ablation_na} show that increasing the number of training samples in either \texttt{Domain} \texttt{A} or \texttt{Domain} \texttt{B} yields enhanced robustness. They are in accordance with the theoretical results in \cite{schmidt_adversarial_robust_data_nips_2018}, which postulate that training a robust classifier requires more data. 

$\bullet$ Fine-tuning improves the Lipschitzness of the loss landscape and hence makes the model more robust. We visualize the histograms of the gradient norms $ \lVert \nabla_{\mathbf{x}} \ell (y, f(\mathbf{x})) \rVert_{2}$ in Fig. \ref{fig:ablation_grad_norms}. Fig. \ref{fig:chist_gn} shows that the gradient norms of the \texttt{FT} model is more likely to have a small value while the maximum value of the gradient norms of the \texttt{Scratch} model can be larger than $2$. The histograms of the gradient norms of the \texttt{Scratch} and \texttt{FT} models are shown in Figs. \ref{fig:scratch_hist_gn} and \ref{fig:ft_hist_gn}, respectively. The adversarial examples that successfully fool the target model are more likely to have large gradient norm values. The gradient norms of the \texttt{FT} model is suppressed, which might improve model robustness. 

\section{Conclusion}
\label{sec:conclusion}
Though fine-tuning is a successful and popular transfer learning technique, its effect on model robustness has been almost ignored. To figure out this problem, extensive experiments are conducted in this paper. The results show that fine-tuning can enhance model robustness under white-box FGSM attacks. We also propose a simple and effective black-box attack method for transfer learning models. Results suggest that fine-tuning might introduce potential risks since a fine-tuned model is more likely to be successfully attacked by the adversarial examples crafted from its source model than a model that is learned from scratch. Our study convinces another advantage of fine-tuning and reveals that there are underlying risks that have been overlooked. We hope that the findings can serve as step stones towards transfer learning models that are both robust and effective. 
In addition, we also developed a new evaluation metric to measure how transferable are the produced adversarial examples to attack transfer learning models. We also believe this metric will be useful for future study of the vulnerability of transfer learning models. 

\begin{acks}
This paper was supported by the Early Career Scheme (ECS, No. 26206717), the Research Impact Fund (RIF, No. R6020-19), the General Research Fund (GRF, No. 16209715 and No. 16244616) from the Research Grants Council (RGC) of Hong Kong, China, and national project 2018AAA0101100. 
\end{acks}

\bibliographystyle{ACM-Reference-Format}
\bibliography{sample-base}


\begin{thebibliography}{27}


\ifx \showCODEN    \undefined \def \showCODEN     #1{\unskip}     \fi
\ifx \showDOI      \undefined \def \showDOI       #1{#1}\fi
\ifx \showISBNx    \undefined \def \showISBNx     #1{\unskip}     \fi
\ifx \showISBNxiii \undefined \def \showISBNxiii  #1{\unskip}     \fi
\ifx \showISSN     \undefined \def \showISSN      #1{\unskip}     \fi
\ifx \showLCCN     \undefined \def \showLCCN      #1{\unskip}     \fi
\ifx \shownote     \undefined \def \shownote      #1{#1}          \fi
\ifx \showarticletitle \undefined \def \showarticletitle #1{#1}   \fi
\ifx \showURL      \undefined \def \showURL       {\relax}        \fi
\providecommand\bibfield[2]{#2}
\providecommand\bibinfo[2]{#2}
\providecommand\natexlab[1]{#1}
\providecommand\showeprint[2][]{arXiv:#2}

\bibitem[\protect\citeauthoryear{Chen, Zhang, Sharma, Yi, and Hsieh}{Chen
  et~al\mbox{.}}{2017}]%
        {chen_zoo_acm_ais_2017}
\bibfield{author}{\bibinfo{person}{Pin-Yu Chen}, \bibinfo{person}{Huan Zhang},
  \bibinfo{person}{Yash Sharma}, \bibinfo{person}{Jinfeng Yi}, {and}
  \bibinfo{person}{Cho-Jui Hsieh}.} \bibinfo{year}{2017}\natexlab{}.
\newblock \showarticletitle{Zoo: Zeroth order optimization based black-box
  attacks to deep neural networks without training substitute models}. In
  \bibinfo{booktitle}{\emph{Proceedings of the 10th ACM Workshop on Artificial
  Intelligence and Security}}. \bibinfo{pages}{15--26}.
\newblock


\bibitem[\protect\citeauthoryear{Chrabaszcz, Loshchilov, and Hutter}{Chrabaszcz
  et~al\mbox{.}}{2017}]%
        {chrabaszcz_imagenet32_arxiv_2017}
\bibfield{author}{\bibinfo{person}{Patryk Chrabaszcz}, \bibinfo{person}{Ilya
  Loshchilov}, {and} \bibinfo{person}{Frank Hutter}.}
  \bibinfo{year}{2017}\natexlab{}.
\newblock \showarticletitle{A downsampled variant of imagenet as an alternative
  to the cifar datasets}.
\newblock \bibinfo{journal}{\emph{arXiv preprint arXiv:1707.08819}}
  (\bibinfo{year}{2017}).
\newblock


\bibitem[\protect\citeauthoryear{Coates, Ng, and Lee}{Coates
  et~al\mbox{.}}{2011}]%
        {coates_stl_aistat_2011}
\bibfield{author}{\bibinfo{person}{Adam Coates}, \bibinfo{person}{Andrew Ng},
  {and} \bibinfo{person}{Honglak Lee}.} \bibinfo{year}{2011}\natexlab{}.
\newblock \showarticletitle{An analysis of single-layer networks in
  unsupervised feature learning}. In \bibinfo{booktitle}{\emph{Proceedings of
  the fourteenth international conference on artificial intelligence and
  statistics}}. \bibinfo{pages}{215--223}.
\newblock


\bibitem[\protect\citeauthoryear{Ganin, Ustinova, Ajakan, Germain, Larochelle,
  Laviolette, Marchand, and Lempitsky}{Ganin et~al\mbox{.}}{2016}]%
        {ganin_dann_jmlr_2016}
\bibfield{author}{\bibinfo{person}{Yaroslav Ganin}, \bibinfo{person}{Evgeniya
  Ustinova}, \bibinfo{person}{Hana Ajakan}, \bibinfo{person}{Pascal Germain},
  \bibinfo{person}{Hugo Larochelle}, \bibinfo{person}{Fran{\c{c}}ois
  Laviolette}, \bibinfo{person}{Mario Marchand}, {and} \bibinfo{person}{Victor
  Lempitsky}.} \bibinfo{year}{2016}\natexlab{}.
\newblock \showarticletitle{Domain-adversarial training of neural networks}.
\newblock \bibinfo{journal}{\emph{The Journal of Machine Learning Research}}
  \bibinfo{volume}{17}, \bibinfo{number}{1} (\bibinfo{year}{2016}),
  \bibinfo{pages}{2096--2030}.
\newblock


\bibitem[\protect\citeauthoryear{Girshick, Donahue, Darrell, and
  Malik}{Girshick et~al\mbox{.}}{2014}]%
        {girshick_cvpr_2014}
\bibfield{author}{\bibinfo{person}{Ross Girshick}, \bibinfo{person}{Jeff
  Donahue}, \bibinfo{person}{Trevor Darrell}, {and} \bibinfo{person}{Jitendra
  Malik}.} \bibinfo{year}{2014}\natexlab{}.
\newblock \showarticletitle{Rich feature hierarchies for accurate object
  detection and semantic segmentation}. In
  \bibinfo{booktitle}{\emph{Proceedings of the IEEE conference on Computer
  Vision and Pattern Recognition}}. \bibinfo{pages}{580--587}.
\newblock


\bibitem[\protect\citeauthoryear{Goodfellow, Shlens, and Szegedy}{Goodfellow
  et~al\mbox{.}}{2015}]%
        {goodfellow_fgsm_iclr_2014}
\bibfield{author}{\bibinfo{person}{Ian Goodfellow}, \bibinfo{person}{Jonathon
  Shlens}, {and} \bibinfo{person}{Christian Szegedy}.}
  \bibinfo{year}{2015}\natexlab{}.
\newblock \showarticletitle{Explaining and harnessing adversarial examples}.
\newblock  (\bibinfo{year}{2015}).
\newblock
\urldef\tempurl%
\url{http://arxiv.org/abs/1412.6572}
\showURL{%
\tempurl}


\bibitem[\protect\citeauthoryear{Hendrycks, Lee, and Mazeika}{Hendrycks
  et~al\mbox{.}}{2019}]%
        {hendrycks_pretrain_robustness_icml_2019}
\bibfield{author}{\bibinfo{person}{Dan Hendrycks}, \bibinfo{person}{Kimin Lee},
  {and} \bibinfo{person}{Mantas Mazeika}.} \bibinfo{year}{2019}\natexlab{}.
\newblock \showarticletitle{Using Pre-Training Can Improve Model Robustness and
  Uncertainty}. In \bibinfo{booktitle}{\emph{International Conference on
  Machine Learning}}. \bibinfo{pages}{2712--2721}.
\newblock


\bibitem[\protect\citeauthoryear{Hull}{Hull}{1994}]%
        {hull_usps_TPAMI_1994}
\bibfield{author}{\bibinfo{person}{Jonathan~J. Hull}.}
  \bibinfo{year}{1994}\natexlab{}.
\newblock \showarticletitle{A database for handwritten text recognition
  research}.
\newblock \bibinfo{journal}{\emph{IEEE Transactions on Pattern Analysis and
  Machine Intelligence}} \bibinfo{volume}{16}, \bibinfo{number}{5}
  (\bibinfo{year}{1994}), \bibinfo{pages}{550--554}.
\newblock


\bibitem[\protect\citeauthoryear{Karpathy, Toderici, Shetty, Leung, Sukthankar,
  and Fei-Fei}{Karpathy et~al\mbox{.}}{2014}]%
        {karpathy_cvpr_2014}
\bibfield{author}{\bibinfo{person}{Andrej Karpathy}, \bibinfo{person}{George
  Toderici}, \bibinfo{person}{Sanketh Shetty}, \bibinfo{person}{Thomas Leung},
  \bibinfo{person}{Rahul Sukthankar}, {and} \bibinfo{person}{Li Fei-Fei}.}
  \bibinfo{year}{2014}\natexlab{}.
\newblock \showarticletitle{Large-scale video classification with convolutional
  neural networks}. In \bibinfo{booktitle}{\emph{Proceedings of the IEEE
  conference on Computer Vision and Pattern Recognition}}.
  \bibinfo{pages}{1725--1732}.
\newblock


\bibitem[\protect\citeauthoryear{Krizhevsky, Hinton, et~al\mbox{.}}{Krizhevsky
  et~al\mbox{.}}{2009}]%
        {krizhevsky_cifar_tr_2009}
\bibfield{author}{\bibinfo{person}{Alex Krizhevsky}, \bibinfo{person}{Geoffrey
  Hinton}, {et~al\mbox{.}}} \bibinfo{year}{2009}\natexlab{}.
\newblock \bibinfo{booktitle}{\emph{Learning multiple layers of features from
  tiny images}}.
\newblock \bibinfo{type}{{T}echnical {R}eport}.
  \bibinfo{institution}{Citeseer}.
\newblock


\bibitem[\protect\citeauthoryear{LeCun, Bottou, Bengio, Haffner,
  et~al\mbox{.}}{LeCun et~al\mbox{.}}{1998}]%
        {lecun_mnist_ieee_proc_1998}
\bibfield{author}{\bibinfo{person}{Yann LeCun}, \bibinfo{person}{L{\'e}on
  Bottou}, \bibinfo{person}{Yoshua Bengio}, \bibinfo{person}{Patrick Haffner},
  {et~al\mbox{.}}} \bibinfo{year}{1998}\natexlab{}.
\newblock \showarticletitle{Gradient-based learning applied to document
  recognition}.
\newblock \bibinfo{journal}{\emph{Proc. IEEE}} \bibinfo{volume}{86},
  \bibinfo{number}{11} (\bibinfo{year}{1998}), \bibinfo{pages}{2278--2324}.
\newblock


\bibitem[\protect\citeauthoryear{Liu, Chen, Liu, and Song}{Liu
  et~al\mbox{.}}{2017}]%
        {liu_arch_transfer_iclr_2016}
\bibfield{author}{\bibinfo{person}{Yanpei Liu}, \bibinfo{person}{Xinyun Chen},
  \bibinfo{person}{Chang Liu}, {and} \bibinfo{person}{Dawn Song}.}
  \bibinfo{year}{2017}\natexlab{}.
\newblock \showarticletitle{Delving into transferable adversarial examples and
  black-box attacks}.
\newblock  (\bibinfo{year}{2017}).
\newblock
\urldef\tempurl%
\url{https://openreview.net/forum?id=Sys6GJqxl}
\showURL{%
\tempurl}


\bibitem[\protect\citeauthoryear{Madry, Makelov, Schmidt, Tsipras, and
  Vladu}{Madry et~al\mbox{.}}{2018}]%
        {madry_pgd_iclr_2018}
\bibfield{author}{\bibinfo{person}{Aleksander Madry},
  \bibinfo{person}{Aleksandar Makelov}, \bibinfo{person}{Ludwig Schmidt},
  \bibinfo{person}{Dimitris Tsipras}, {and} \bibinfo{person}{Adrian Vladu}.}
  \bibinfo{year}{2018}\natexlab{}.
\newblock \showarticletitle{Towards deep learning models resistant to
  adversarial attacks}.
\newblock  (\bibinfo{year}{2018}).
\newblock
\urldef\tempurl%
\url{https://openreview.net/forum?id=rJzIBfZAb}
\showURL{%
\tempurl}


\bibitem[\protect\citeauthoryear{Netzer, Wang, Coates, Bissacco, Wu, and
  Ng}{Netzer et~al\mbox{.}}{2011}]%
        {netzer_svhn_tr_2011}
\bibfield{author}{\bibinfo{person}{Yuval Netzer}, \bibinfo{person}{Tao Wang},
  \bibinfo{person}{Adam Coates}, \bibinfo{person}{Alessandro Bissacco},
  \bibinfo{person}{Bo Wu}, {and} \bibinfo{person}{Andrew~Y Ng}.}
  \bibinfo{year}{2011}\natexlab{}.
\newblock \showarticletitle{Reading digits in natural images with unsupervised
  feature learning}.
\newblock  (\bibinfo{year}{2011}).
\newblock


\bibitem[\protect\citeauthoryear{Oquab, Bottou, Laptev, and Sivic}{Oquab
  et~al\mbox{.}}{2014}]%
        {oquab_transfer_midlevel_cvpr_2014}
\bibfield{author}{\bibinfo{person}{Maxime Oquab}, \bibinfo{person}{Leon
  Bottou}, \bibinfo{person}{Ivan Laptev}, {and} \bibinfo{person}{Josef Sivic}.}
  \bibinfo{year}{2014}\natexlab{}.
\newblock \showarticletitle{Learning and transferring mid-level image
  representations using convolutional neural networks}. In
  \bibinfo{booktitle}{\emph{Proceedings of the IEEE conference on Computer
  Vision and Pattern Recognition}}. \bibinfo{pages}{1717--1724}.
\newblock


\bibitem[\protect\citeauthoryear{Pan and Yang}{Pan and Yang}{2010}]%
        {pan2010survey}
\bibfield{author}{\bibinfo{person}{Sinno~Jialin Pan} {and}
  \bibinfo{person}{Qiang Yang}.} \bibinfo{year}{2010}\natexlab{}.
\newblock \showarticletitle{A survey on transfer learning}.
\newblock \bibinfo{journal}{\emph{IEEE Transactions on Knowledge and Data
  Engineering}} \bibinfo{volume}{22}, \bibinfo{number}{10}
  (\bibinfo{year}{2010}), \bibinfo{pages}{1345--1359}.
\newblock


\bibitem[\protect\citeauthoryear{Papernot, McDaniel, Goodfellow, Jha, Celik,
  and Swami}{Papernot et~al\mbox{.}}{2017}]%
        {papernot_substitute_acm_ccs_2017}
\bibfield{author}{\bibinfo{person}{Nicolas Papernot}, \bibinfo{person}{Patrick
  McDaniel}, \bibinfo{person}{Ian Goodfellow}, \bibinfo{person}{Somesh Jha},
  \bibinfo{person}{Z~Berkay Celik}, {and} \bibinfo{person}{Ananthram Swami}.}
  \bibinfo{year}{2017}\natexlab{}.
\newblock \showarticletitle{Practical black-box attacks against machine
  learning}. In \bibinfo{booktitle}{\emph{Proceedings of the 2017 ACM on Asia
  conference on computer and communications security}}.
  \bibinfo{pages}{506--519}.
\newblock


\bibitem[\protect\citeauthoryear{Russakovsky, Deng, Su, Krause, Satheesh, Ma,
  Huang, Karpathy, Khosla, Bernstein, Berg, and Fei-Fei}{Russakovsky
  et~al\mbox{.}}{2015}]%
        {russ_imagenet_ijcv_2015}
\bibfield{author}{\bibinfo{person}{Olga Russakovsky}, \bibinfo{person}{Jia
  Deng}, \bibinfo{person}{Hao Su}, \bibinfo{person}{Jonathan Krause},
  \bibinfo{person}{Sanjeev Satheesh}, \bibinfo{person}{Sean Ma},
  \bibinfo{person}{Zhiheng Huang}, \bibinfo{person}{Andrej Karpathy},
  \bibinfo{person}{Aditya Khosla}, \bibinfo{person}{Michael Bernstein},
  \bibinfo{person}{Alexander~C. Berg}, {and} \bibinfo{person}{Li Fei-Fei}.}
  \bibinfo{year}{2015}\natexlab{}.
\newblock \showarticletitle{{ImageNet Large Scale Visual Recognition
  Challenge}}.
\newblock \bibinfo{journal}{\emph{International Journal of Computer Vision}}
  \bibinfo{volume}{115}, \bibinfo{number}{3} (\bibinfo{year}{2015}),
  \bibinfo{pages}{211--252}.
\newblock


\bibitem[\protect\citeauthoryear{Schmidt, Santurkar, Tsipras, Talwar, and
  Madry}{Schmidt et~al\mbox{.}}{2018}]%
        {schmidt_adversarial_robust_data_nips_2018}
\bibfield{author}{\bibinfo{person}{Ludwig Schmidt}, \bibinfo{person}{Shibani
  Santurkar}, \bibinfo{person}{Dimitris Tsipras}, \bibinfo{person}{Kunal
  Talwar}, {and} \bibinfo{person}{Aleksander Madry}.}
  \bibinfo{year}{2018}\natexlab{}.
\newblock \showarticletitle{Adversarially robust generalization requires more
  data}. In \bibinfo{booktitle}{\emph{Advances in Neural Information Processing
  Systems}}. \bibinfo{pages}{5014--5026}.
\newblock


\bibitem[\protect\citeauthoryear{Szegedy, Zaremba, Sutskever, Bruna, Erhan,
  Goodfellow, and Fergus}{Szegedy et~al\mbox{.}}{2014}]%
        {szegedy_iclr_2013}
\bibfield{author}{\bibinfo{person}{Christian Szegedy},
  \bibinfo{person}{Wojciech Zaremba}, \bibinfo{person}{Ilya Sutskever},
  \bibinfo{person}{Joan Bruna}, \bibinfo{person}{Dumitru Erhan},
  \bibinfo{person}{Ian Goodfellow}, {and} \bibinfo{person}{Rob Fergus}.}
  \bibinfo{year}{2014}\natexlab{}.
\newblock \showarticletitle{Intriguing properties of neural networks}.
\newblock  (\bibinfo{year}{2014}).
\newblock
\urldef\tempurl%
\url{http://arxiv.org/abs/1312.6199}
\showURL{%
\tempurl}


\bibitem[\protect\citeauthoryear{Venugopalan, Rohrbach, Donahue, Mooney,
  Darrell, and Saenko}{Venugopalan et~al\mbox{.}}{2015a}]%
        {venugopalan_iccv_2015}
\bibfield{author}{\bibinfo{person}{Subhashini Venugopalan},
  \bibinfo{person}{Marcus Rohrbach}, \bibinfo{person}{Jeffrey Donahue},
  \bibinfo{person}{Raymond Mooney}, \bibinfo{person}{Trevor Darrell}, {and}
  \bibinfo{person}{Kate Saenko}.} \bibinfo{year}{2015}\natexlab{a}.
\newblock \showarticletitle{Sequence to sequence-video to text}. In
  \bibinfo{booktitle}{\emph{Proceedings of the IEEE conference on Computer
  Vision and Pattern Recognition}}. \bibinfo{pages}{4534--4542}.
\newblock


\bibitem[\protect\citeauthoryear{Venugopalan, Xu, Donahue, Rohrbach, Mooney,
  and Saenko}{Venugopalan et~al\mbox{.}}{2015b}]%
        {venugopalan_naacl_2015}
\bibfield{author}{\bibinfo{person}{Subhashini Venugopalan},
  \bibinfo{person}{Huijuan Xu}, \bibinfo{person}{Jeff Donahue},
  \bibinfo{person}{Marcus Rohrbach}, \bibinfo{person}{Raymond Mooney}, {and}
  \bibinfo{person}{Kate Saenko}.} \bibinfo{year}{2015}\natexlab{b}.
\newblock \showarticletitle{Translating Videos to Natural Language Using Deep
  Recurrent Neural Networks}. In \bibinfo{booktitle}{\emph{Proceedings of the
  2015 Conference of the North American Chapter of the Association for
  Computational Linguistics: Human Language Technologies}}.
  \bibinfo{pages}{1494--1504}.
\newblock


\bibitem[\protect\citeauthoryear{Vinyals, Toshev, Bengio, and Erhan}{Vinyals
  et~al\mbox{.}}{2015}]%
        {vinyals_cvpr_2015}
\bibfield{author}{\bibinfo{person}{Oriol Vinyals}, \bibinfo{person}{Alexander
  Toshev}, \bibinfo{person}{Samy Bengio}, {and} \bibinfo{person}{Dumitru
  Erhan}.} \bibinfo{year}{2015}\natexlab{}.
\newblock \showarticletitle{Show and tell: A neural image caption generator}.
  In \bibinfo{booktitle}{\emph{Proceedings of the IEEE conference on Computer
  Vision and Pattern Recognition}}. \bibinfo{pages}{3156--3164}.
\newblock


\bibitem[\protect\citeauthoryear{Weiss, Khoshgoftaar, and Wang}{Weiss
  et~al\mbox{.}}{2016}]%
        {weiss2016survey}
\bibfield{author}{\bibinfo{person}{Karl Weiss}, \bibinfo{person}{Taghi~M
  Khoshgoftaar}, {and} \bibinfo{person}{Dingding Wang}.}
  \bibinfo{year}{2016}\natexlab{}.
\newblock \showarticletitle{A survey of transfer learning}.
\newblock \bibinfo{journal}{\emph{Journal of Big Data}} \bibinfo{volume}{3},
  \bibinfo{number}{1} (\bibinfo{year}{2016}), \bibinfo{pages}{1--40}.
\newblock


\bibitem[\protect\citeauthoryear{Yosinski, Clune, Bengio, and Lipson}{Yosinski
  et~al\mbox{.}}{2014}]%
        {yosinski_nips_2014}
\bibfield{author}{\bibinfo{person}{Jason Yosinski}, \bibinfo{person}{Jeff
  Clune}, \bibinfo{person}{Yoshua Bengio}, {and} \bibinfo{person}{Hod Lipson}.}
  \bibinfo{year}{2014}\natexlab{}.
\newblock \showarticletitle{How transferable are features in deep neural
  networks?}. In \bibinfo{booktitle}{\emph{Advances in Neural Information
  Processing Systems}}. \bibinfo{pages}{3320--3328}.
\newblock


\bibitem[\protect\citeauthoryear{Yuan, He, Zhu, and Li}{Yuan
  et~al\mbox{.}}{2019}]%
        {yuan_tnnls_attac_survey_2019}
\bibfield{author}{\bibinfo{person}{Xiaoyong Yuan}, \bibinfo{person}{Pan He},
  \bibinfo{person}{Qile Zhu}, {and} \bibinfo{person}{Xiaolin Li}.}
  \bibinfo{year}{2019}\natexlab{}.
\newblock \showarticletitle{Adversarial examples: Attacks and defenses for deep
  learning}.
\newblock \bibinfo{journal}{\emph{IEEE Transactions on Neural Networks and
  Learning Systems}} \bibinfo{volume}{30}, \bibinfo{number}{9}
  (\bibinfo{year}{2019}), \bibinfo{pages}{2805--2824}.
\newblock


\bibitem[\protect\citeauthoryear{Zagoruyko and Komodakis}{Zagoruyko and
  Komodakis}{2016}]%
        {zagoruyko_wrn_arxiv_2016}
\bibfield{author}{\bibinfo{person}{Sergey Zagoruyko} {and}
  \bibinfo{person}{Nikos Komodakis}.} \bibinfo{year}{2016}\natexlab{}.
\newblock \showarticletitle{Wide residual networks}.
\newblock \bibinfo{journal}{\emph{arXiv preprint arXiv:1605.07146}}
  (\bibinfo{year}{2016}).
\newblock


\end{thebibliography}



\end{document}